\documentclass[letterpaper, 10 pt, conference]{ieeeconf}
\IEEEoverridecommandlockouts
\IEEEoverridecommandlockouts
\usepackage{cite}
\usepackage{amsmath,amssymb,amsfonts}
\usepackage{algorithm}
\usepackage{algpseudocode}
\usepackage{graphicx}
\usepackage{subfigure}
\usepackage{textcomp}
\usepackage{multirow}
\usepackage{diagbox}
\usepackage{hyperref}
\usepackage{mdframed}
\usepackage{caption}
\usepackage{lineno}
\usepackage{needspace}
\hypersetup{
    colorlinks=true,
    linkcolor=blue,
    filecolor=cyan,      
    urlcolor=magenta,
}
\usepackage{subcaption}
\usepackage{listings}
\lstalias{plantuml}{}
\usepackage{xcolor}
\usepackage{lipsum}
\usepackage[most]{tcolorbox}
\usepackage{listings}
\usepackage{xcolor}
\definecolor{lightblue}{rgb}{0.68, 0.85, 0.9} 
\definecolor{lightgreen}{rgb}{0.85, 0.93, 0.85}
\usepackage{tikz}

\usepackage[margin=2cm]{geometry}
\usepackage[most]{tcolorbox}
\newtcolorbox{mytextbox}[1][]{%
  sharp corners,
  enhanced,
  colback=white,
  colframe=lightgray,
  boxrule=0.5pt, 
  left=1mm,
  before skip=1.5pt, 
  after skip=1.5pt, 
  fontupper=\fontsize{8.5pt}{11pt}\selectfont, 
  fontlower=\fontsize{8.5pt}{11pt}\selectfont, 
}

\newtcolorbox{grd_format}[1][]{%
  sharp corners,
  enhanced,
  colback=white,
  colframe=lightgray,
  boxrule=1.5pt,
  attach title to upper,
  left=1mm,
  fontupper=\fontsize{9pt}{9pt}\selectfont\linespread{1.5}\selectfont, 
  fontlower=\fontsize{10pt}{12pt}\selectfont,
  width=\linewidth,
  lineskip=7pt,
  #1
}

\newtcolorbox{grd_output}[1][]{%
  sharp corners,
  enhanced,
  colback=white,
  colframe=lightgray,
  boxrule=0.5pt, 
  attach title to upper,
  left=0mm,
  fontupper=\fontsize{9pt}{9pt}\selectfont, 
  fontlower=\fontsize{10pt}{12pt}\selectfont, 
  width=4.5cm,
  lineskip=3pt
}

\newtcolorbox{prd_format}[1][]{%
  sharp corners,
  enhanced,
  colback=white,
  colframe=lightgray,
  boxrule=1.5pt,
  attach title to upper,
  left=1mm,
  fontupper=\fontsize{9pt}{9pt}\selectfont\linespread{1.5}\selectfont, 
  fontlower=\fontsize{10pt}{12pt}\selectfont,
  width=\linewidth,
  lineskip=7pt,
  #1
}

\usepackage{pythonhighlight}
\definecolor{bpink}{rgb}{0.99, 0.84, 0.79}
\definecolor{bubbles}{rgb}{0.91, 1.0, 1.0}
\definecolor{champagne}{rgb}{0.97, 0.91, 0.81}

\def\BibTeX{{\rm B\kern-.05em{\sc i\kern-.025em b}\kern-.08em
    T\kern-.1667em\lower.7ex\hbox{E}\kern-.125emX}}
\begin{document}

\title{\LARGE \bf
Visual Preference Inference: An Image Sequence-Based Preference Reasoning in Tabletop Object Manipulation
}

\author{{Joonhyung~Lee$^{1}$,
        Sangbeom~Park$^{1}$,
        Yongin Kwon$^{2}$,
        Jemin Lee$^{2}$,
        Minwook Ahn$^{3}$
        and~Sungjoon~Choi$^{1*}$}
\thanks{This work was supported by Institute of Information \& communications Technology Planning \& Evaluation (IITP) grant funded by the Korea government (MSIT) (No. 2019-0-00079 , Artificial Intelligence Graduate School Program (Korea University)), and Institute of Information \& communications Technology Planning \& Evaluation (IITP) grant funded by the Korea government(MSIT) (No.RS-2023-00277060, Development of open edge AI SoC hardware and software platform.)}
\thanks{$^{1}$Joonhyung Lee, Sangbeom~Park, and Sungjoon~Choi are with the Department of Artificial Intelligence, Korea University, Seoul, Korea
(email: {\tt\small \{dlwnsgud8823, sangbeom-park, sungjoon-choi\}@korea.ac.kr})}%
\thanks{$^{2}$Yongin Kwon, and Jemin Lee are with the Electronics and Telecommunications Research Institute, Daejeon, Korea
(email: {\tt\small \{yongin.kwon,~leejaymin\}@etri.re.kr})} %
\thanks{$^{3}$Minwook Ahn is with the Neubla, Seoul, Korea
(email: {\tt\small minwook.ahn@neubla.com})} %
\thanks{* Corresponding Author} %
}
\maketitle

%
%
\begin{abstract}
In robotic object manipulation, human preferences can often be influenced by the visual attributes of objects, such as color and shape. 
These properties play a crucial role in operating a robot to interact with objects and align with human intention.
In this paper, we focus on the problem of inferring underlying human preferences from a sequence of raw visual observations in tabletop manipulation environments with a variety of object types, named Visual Preference Inference (VPI).
To facilitate visual reasoning in the context of manipulation, we introduce the Chain-of-Visual-Residuals (CoVR) method. CoVR employs a prompting mechanism that describes the difference between the consecutive images (i.e., visual residuals) and incorporates such texts with a sequence of images to infer the user's preference.
This approach significantly enhances the ability to understand and adapt to dynamic changes in its visual environment during manipulation tasks. Furthermore, we incorporate such texts along with a sequence of images to infer the user's preferences. 
Our method outperforms baseline methods in terms of extracting human preferences from visual sequences in both simulation and real-world environments.
Code and videos are available at: \href{https://joonhyung-lee.github.io/vpi/}{https://joonhyung-lee.github.io/vpi/}
\end{abstract}


%
%
\section{Introduction}
Recent research has actively focused on aligning the behaviors of a robot or AI systems to match user preferences, thereby enhancing interaction and task performance efficiency~\cite{ouyang22instructgpt}.
For example, when a robotic arm performs the task of pouring the liquid inside a cup into a bowl, the trajectory planning can be optimized by aligning with a user's intuitive control preferences~\cite{li_20_learning_user_preference}.
Commonly, robotic behaviors have mainly relied on manually designed features through scalar values, such as walking motions for quadruped robots~\cite{lee_2021_pebble}, velocity for autonomous driving~\cite{wilde_20_active_preference}, and gait patterns for exoskeletons~\cite{lee_23_exo}. 
However, these approaches are limited in that preferences have yet to be extended to visual features from images that enable capturing the context of the current scene intuitively.

Multimodal large language models (MLLMs) have advanced by integrating direct sensory perception into the reasoning processes, enhancing the ability to interpret and generate human-like responses~\cite{liu_23_llava, 00_GPT_4V, zhao_23_lavila}.
These models have the ability to adapt contextually based on visual observations and respond to language descriptions appropriately.
Furthermore, MLLMs have achieved human-like reasoning performances in a variety of robotic tasks (e.g., understanding complex multi-object relations \cite{kapelyukh_23_dream2real} and long-horizon planning in manipulation tasks \cite{hu_23_vila}). 
Human-like reasoning based on visual information has great potential as it can predict human preferences without manually defining features.
In this work, our goal is to extract human preferences from raw visual information such as semantic (e.g., color and shape) or spatial (e.g., arrangement pattern) features.

\begin{figure}[t!]
    \centering
    \includegraphics[width=0.49\textwidth]{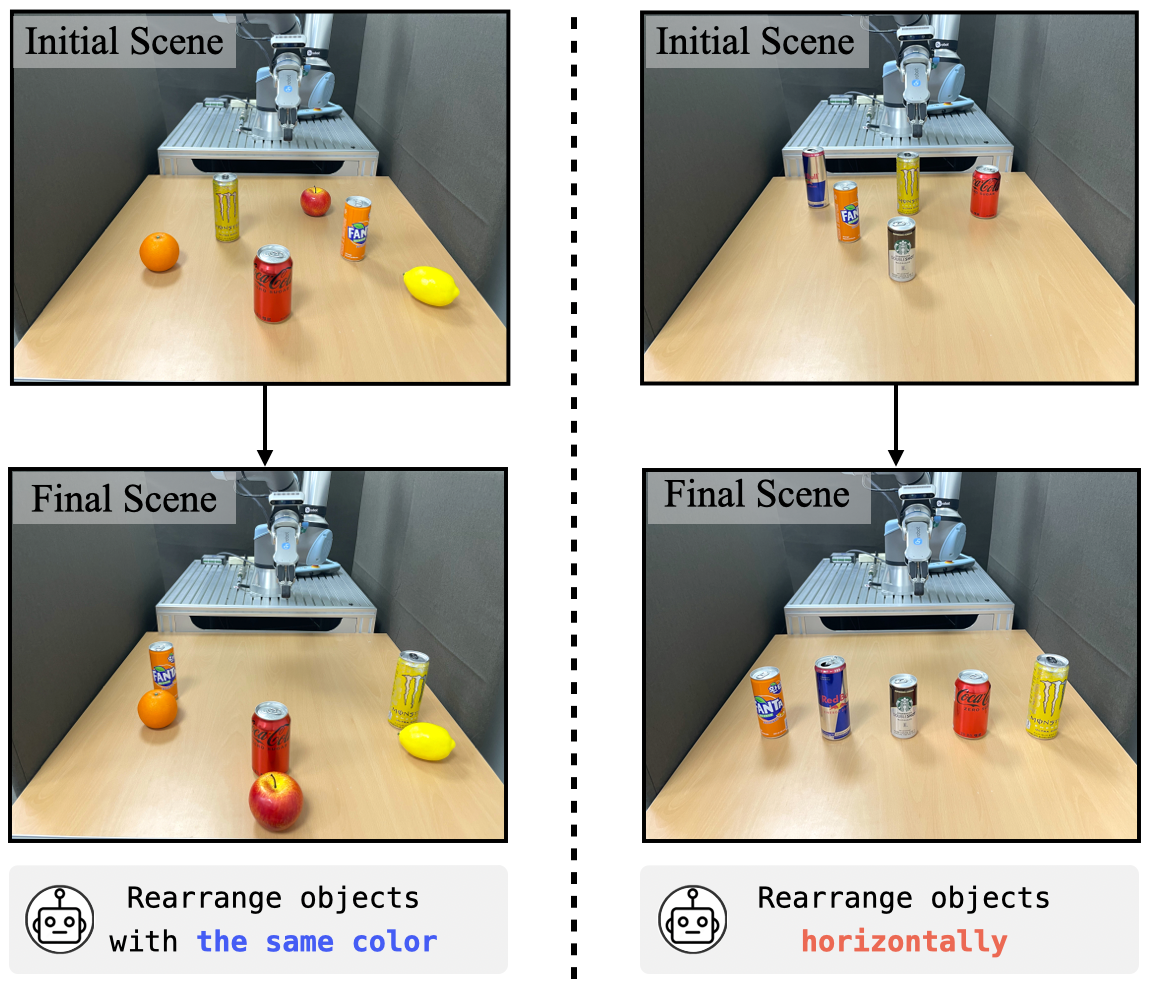}
        \caption{\textbf{Visual Preference Inference (VPI) Tasks.} We define VPI tasks as reasoning user preferences based on an image sequence. 
        Specifically, the task involves a robot that moves objects to target locations, following user instructions via mouse clicks which provide which object to move and where to place it.
    }
\label{fig:vpi-concept}
\end{figure}

\begin{figure*}[h!]
    \centering
    \includegraphics[width=0.95\textwidth]{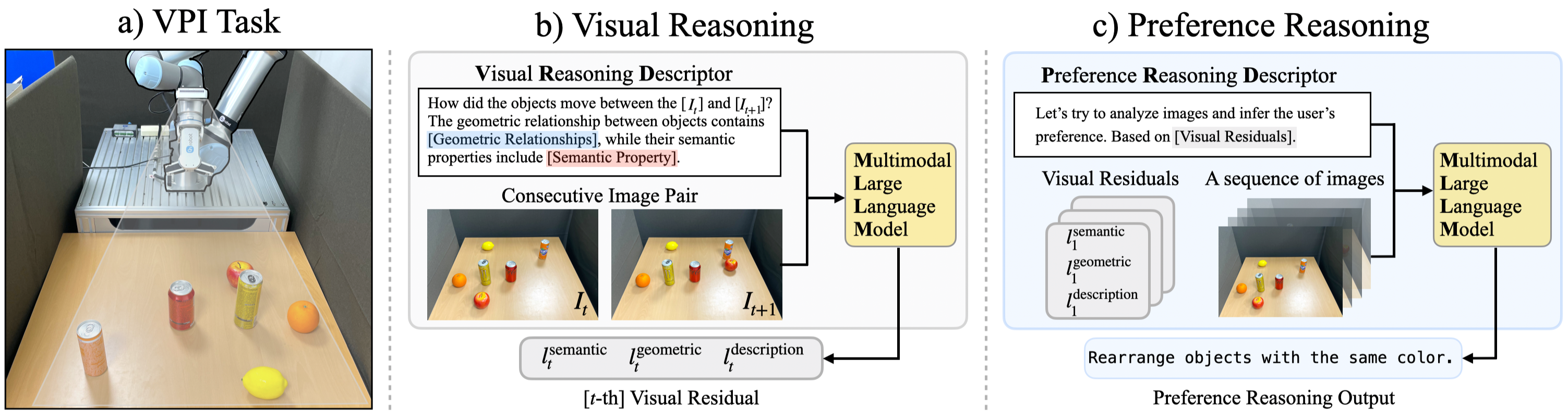}
    \caption{
    \textbf{Overview of Chain-of-Visual-Residuals:} (a) We introduce a Visual Preference Inference (VPI) task, which extracts users' preferences solely from visual representations in tabletop manipulation environments. 
    Our approach, CoVR prompting, involves generating (b) visual reasoning descriptions of consecutive images and (c) chaining these descriptions for interpreting human preferences from the scene sequences.}
\label{fig:overview}
\end{figure*}

In particular, we focus on inferring the human preferences that require visual understanding aligned with the user's intentions within robotic manipulation tasks. 
Hence, we introduce the task of extracting user's preferences solely from visual representations, referred to as $\textbf{VPI}$ which stands for $\textbf{V}$isual $\textbf{P}$reference $\textbf{I}$nference. 
To this end, we propose $\textbf{C}$hain-$\textbf{o}$f-$\textbf{V}$isual-$\textbf{R}$esiduals ($\textbf{CoVR}$) prompting, a method that involves a series of intermediate visual reasoning steps leading to the end response. 
Specifically, $\textbf{CoVR}$ consists of two phases: 1) Visual Reasoning Descriptor (VRD), which maps user interaction with image inputs into scene descriptions that focus on capturing both semantic attributes of objects and changes in the geometric relationships between objects, and 2) Preference Reasoning Descriptor (PRD), which predicts a suitable preference considering the interactions of object manipulation. 

We demonstrate the versatility of our method across a number of object manipulation tasks, as illustrated in Fig.~\ref{fig:vpi-concept}. The results indicate that our method adeptly infers appropriate interaction patterns in response to dynamic changes of moved objects, including semantic reasoning tasks (e.g., the rearrangement of identically colored objects). 
Furthermore, the CoVR prompting is suitable for spatial reasoning tasks, which are notably difficult to articulate using textual descriptions alone (e.g., the horizontal rearrangement of objects).

Our contributions are summarized as follows: First, we present a visual reasoning task for inferring human preferences from a sequence of images.
Second, we propose the CoVR prompting that enables inferring user preferences while understanding the dynamic changes from visual information.
Third, we show that our proposed method can predict diverse user preferences in multiple manipulation tasks with various object types.

%
%
\section{Related Work}
\subsection{Human Preferences in Robotics}

Research on predicting user preferences has been actively conducted in robotics~\cite{wilde_18_pbl_linear,bobu_22_inducing}.
In particular, these approaches have been utilized for a furniture assembly task where the robot assists the human by using a screwdriver on a small screw~\cite{munzer_17_preference_furniture, grigore_18_preference_furniture}. 
In addition, the quadruped robot’s behavior can naturally be tailored to one’s preference, such as waving the left front leg~\cite{lee_2021_pebble}. 
Humanoid motion can also be optimized with respect to the user's preferences~\cite{guan_23_relative, dong_23_aligndiff}.
Similar to the aforementioned approach, preferences can be designed by hand-crafted features.
Moreover, a self-driving car has diverse features, including distance to other vehicles, speed, and heading angle~\cite{biyik_18_batch_active}.
Other studies have explored the exoskeleton domain to extract the gait pattern~\cite{tucker_20_preferencebased, lee_23_exo}. 
Additionally, research on preference prediction has also been extended to using visual features for determining mobile robot path-planning preferences~\cite{sikand_22_visual_pbl}.
However, \cite{sikand_22_visual_pbl} also mentions that this paper is limited to handling the preference features that require interpreting visual information contextually. 
In contrast, we focus on reasoning visual information based on common knowledge for human-like prediction.

\subsection{Multimodal Large models for Robotics}
Large language models have been used for task planning~\cite{brohan_23_saycan}, code generation~\cite{liang_23_code_as_policies}, reward design~\cite{kwon_23_reward_design}, contact scheduling~\cite{tang_23_saytap}, and reasonableness reasoning~\cite{lee_23_spots} in robotics. Moreover, these foundation models can reason the common knowledge for understanding the geometric relationship between objects for placement tasks in the real-world~\cite {ding_2023_llm-grop}. 
However, relying solely on textual descriptions can make it difficult to connect with common knowledge.
In contrast, visual representations make it easier to infer visual attributes, particularly when linking common knowledge to the physical world through sensory inputs.
Accordingly, PG-VLM~\cite{gao_23pg-vlm} leverages a multimodal LLM as a scalable way of providing high-level physical reasoning that humans use to interact with the world. 
Extensive work on enhancing the visual understanding capacity of multimodal LLMs by simply overlaying spatial and speakable marks on the images are mentioned in ~\cite{yang_23_som}. Similarly, 3DAP~\cite{liu_23_3d_prompts} has utilized the visual prompt method to maximize 3D understanding capabilities. 
Inspired by the visual reasoning capabilities of the foundation model, our method uses visual chain-of-thought prompts to infer user preferences.

%
%
\section{Problem Formulation}

In this section, we address the problem of extracting human preferences from visual representations (i.e., a set of RGB images), referred to as \textbf{VPI}: \textbf{V}isual \textbf{P}reference \textbf{I}nference. 
In particular, we focus on tabletop manipulation tasks and formulate VPI to interpret human preferences using a sequence of \( n \) images \( \mathcal{I} = \{I_1, I_2, \cdots, I_n\} \) obtained from a camera mounted on the end effector.

Our method focuses on analyzing visual signals from images to understand both the semantic and geometric properties of the objects in the scene. 
Specifically, the semantic property contains object color, shape, and category, and the geometric property corresponds to inferring the relative positions between objects, displacements, and the initial and final locations of the objects.
As our method does not rely on the manual design of features, such properties are obtained solely from visual information utilizing an MLLM with proper prompting method which will be described in the next section. 

%
%
\section{Proposed Method}
\label{method}

We propose \textbf{C}hain-\textbf{o}f-\textbf{V}isual-\textbf{R}esiduals~(\textbf{CoVR}) prompting, a method that connects visual understandings to reason about preferences from a long-horizon image sequence.
Our proposed approach is comprised of two key components: Visual Reasoning Descriptor (VRD) in Sec.~\ref{visual-reasoning} and Preference Reasoning Descriptor (PRD) in Sec.~\ref{preference-reasoning}.
Specifically, VRD is a template-based prompting method designed to understand semantic properties and identify changes in how the objects are geometrically related in consecutive image pairs.
After the aforementioned process has been completed, PRD connects the previously obtained textual scene descriptions with input images to enhance the effectiveness of predicting a user's intention.
The overall pipeline is depicted in Fig.~\ref{fig:overview}.

\subsection{Visual Reasoning Descriptor}
\label{visual-reasoning}
Our goal is to identify \textit{which} object has moved between two consecutive images and \textit{how} the geometric relationship of objects has changed, while simultaneously inferring the semantic properties of each object.
To this end, we present Visual Reasoning Descriptor (VRD) which translates input images into natural language scene descriptions (referred to as visual residuals). 
Visual residual $V$ contains both the semantic properties of the objects and the difference in the objects' configurations between consecutive image pairs and consists of three components: $\{l^{\text{semantic}}, l^{\text{geometric}}, l^{\text{description}}\}$.
\( l^{\text{semantic}} \) describes the semantic property of two objects (i.e., source object and target object) that have moved in between the image pairs and have been involved in this movement, \( l^{\text{geometric}} \) refers to the spatial arrangement of the resulting relationship between two objects, and \( l^{\text{description}} \) refers to the scene description of consecutive image pairs.
VRD process can be formulated as:
\begin{equation*}
    \text{VRD}(I_{n-1}, I_{n}) \rightarrow V_{n-1}:= \{l^{\text{semantic}}_{n-1}, l^{\text{geometric}}_{n-1}, l^{\text{description}}_{n-1}\},
\end{equation*}
\noindent where \(n\) represent image sequence index and $V_{n}$ describes visual residual for the image pair \( (I_{n-1}, I_{n}) \).

Building upon this formulation, we provide the MLLM with the instructions along with the whole image sequence $\mathcal{I}$.
However, when handling a large number of images, MLLMs tend to suffer from a lack of accuracy in interpreting scene information.
To handle this issue, VRD first extracts the given image sequence into consecutive pairs and computes visual residual $V$ by utilizing few-shot prompting where the prompts are given as follows:\footnote{
Geometric relations and semantic property include \{to the left of, to the right of, in front of, behind of\} and \{color, shape, category\}, respectively. 
}
\begin{mytextbox}
I will give you a set of images \textcolor{gray}{$\text{[image1}$, $\text{image2}$, \ldots, $\text{imageN]}$}. \\
The goal is to reason about the \colorbox{white}{geometric} and semantic properties of objects in an image sequence. \\
Format: \\ 
- \textcolor{red!50}{geometric property}\\
- \textcolor{blue!50}{semantic property}\\
- \textcolor{orange!65}{description} \\
\textcolor{gray}{{\text{[Examples]}}} \\
How did the objects move between the \textcolor{gray}{$\text{[image1]}$} and \textcolor{gray}{$\text{[image2]}$}? \\
The geometric relationship between objects contains \textcolor{gray}{\text{[Geometric Relationships]}}, while their semantic properties include \textcolor{gray}{\text{[Semantic Property]}}.
\end{mytextbox}
\noindent As above, the VRD recognizes both the \textcolor{red!50}{geometric} and \textcolor{blue!50}{semantic} properties of objects and provides a textual scene \textcolor{orange!65}{description} to identify which objects have been moved and the corresponding relationships in between images.

For example in the case of Fig.~\ref{fig:overview}-$(I_{1},I_{2})$, scene description can be prompted by VRD to link object names (``apple", ``orange drink"), semantic attribute (``sphere-shaped"), and geometric relationship (``in front of").
The generated response (\colorbox{green!10}{highlighted}) is as follows:
\begin{mytextbox}
\textcolor{red!50}{geometric property}: \colorbox{green!10}{in\_front\_of} \\
\textcolor{blue!50}{semantic property}: \colorbox{green!10}{source object: apple, red, sphere\_shaped,}\\
\colorbox{green!10}{target object: orange drink, orange, cylinder\_shaped}\\
\textcolor{orange!65}{description}: \colorbox{green!10}{Move the apple in front of the orange drink.}
\end{mytextbox}
\noindent Here, the source object describes the object that has moved in between these image pairs, and the target object refers to the object that has been involved in this movement. We repeat this process iteratively to obtain visual residuals from all the successive image pairs.

\subsection{Preference Reasoning Descriptor}
\label{preference-reasoning}
To interpret the overall preference from the obtained sequence of visual residuals  \( \mathcal{V}=\{V_{1},\cdots,V_{n-1}\} \) between an image sequence \( \mathcal{I} \) of length $n$, we propose Preference Reasoning Descriptor (PRD) to interpret user preferences described in natural language descriptions.
To this end, we propose Preference Reasoning Descriptor (PRD) to interpret user preferences described in natural language descriptions.
The visual residual information (obtained from VRD) along with the original image sequence is fed into PRD to reason about the underlying human preferences. 

Given a set of images \( \mathcal{I} \) and a sequence of visual residuals \( \mathcal{V} \) for each image pair in \( \mathcal{I} \), we formulate PRD that infers preferred objectives within the set of predefined preferences:
\begin{equation*}
    \text{PRD}(\mathcal{I}, \mathcal{V}) \rightarrow l^{\text{preference}},
\end{equation*}
\noindent where \( l^{\text{preference}} \) denotes the inferred preferences based on the visual residual information. 
Specifically, we define a preference set containing nine elements for the few-shot prompting method\footnote{
The whole prompts can be found on \href{https://joonhyung-lee.github.io/vpi/}{https://joonhyung-lee.github.io/vpi/}
}:
\begin{mytextbox}
Rearrange objects with the same color. \\
Group objects by the same shape. \\
Make objects into a horizontal line. \\
Sort objects vertically. \\
\ldots
\end{mytextbox}

Based on the above preference set, PRD infers the user preferences (\colorbox{green!10}{highlighted}) with previously obtained visual residual components (in \textcolor{gray}{gray}):
\begin{mytextbox}
Let's try to analyze images and infer the user's preference. \\
Based on previous visual residuals: \textcolor{gray}{\text{[Visual Residuals]}} \\
Preference: \colorbox{green!10}{Rearrange objects with the same color.}
\end{mytextbox}
We note that our method is capable of inferring human preferences in an open-ended manner without giving a predefined preference set.
However, explicitly giving the preference set is more effective in terms of evaluating the performance of our method and baselines. 

%
%
\section{Experiments}
In this section, we designed our experiments to address the following questions: (1) Can our proposed visual reasoning descriptor (VRD) capture both semantic and geometric properties from images during tabletop manipulation tasks? (2) Can our proposed preference reasoning descriptor (PRD) accurately predict human preferences by utilizing visual residuals extracted from raw observations in multiple manipulation scenarios? 
\begin{figure}[t!] 
	\centering
	\begin{minipage}{\linewidth}
		\centering
		\subfigure[\textbf{Block} Task]{
			\includegraphics[width=0.45\linewidth]{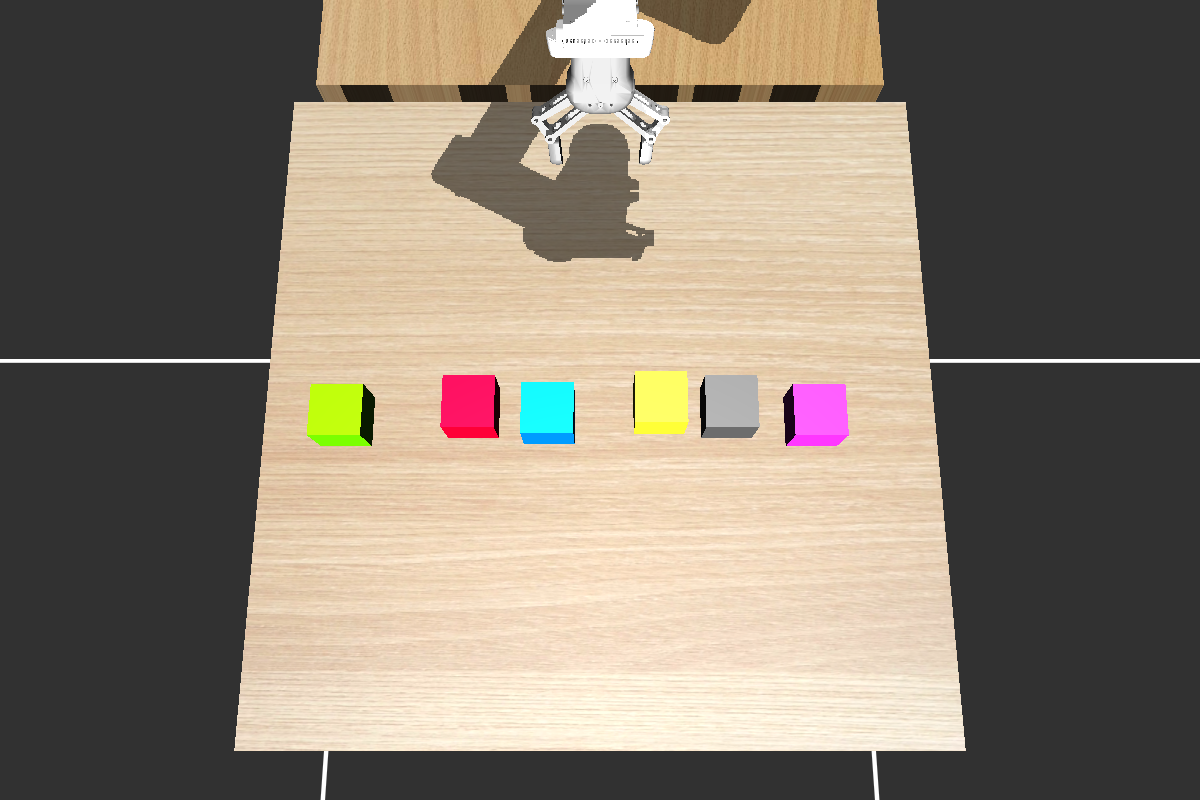}
			\label{blocks-only-preference}
		}
		\subfigure[\textbf{Polygon} Task]{
			\includegraphics[width=0.45\linewidth]{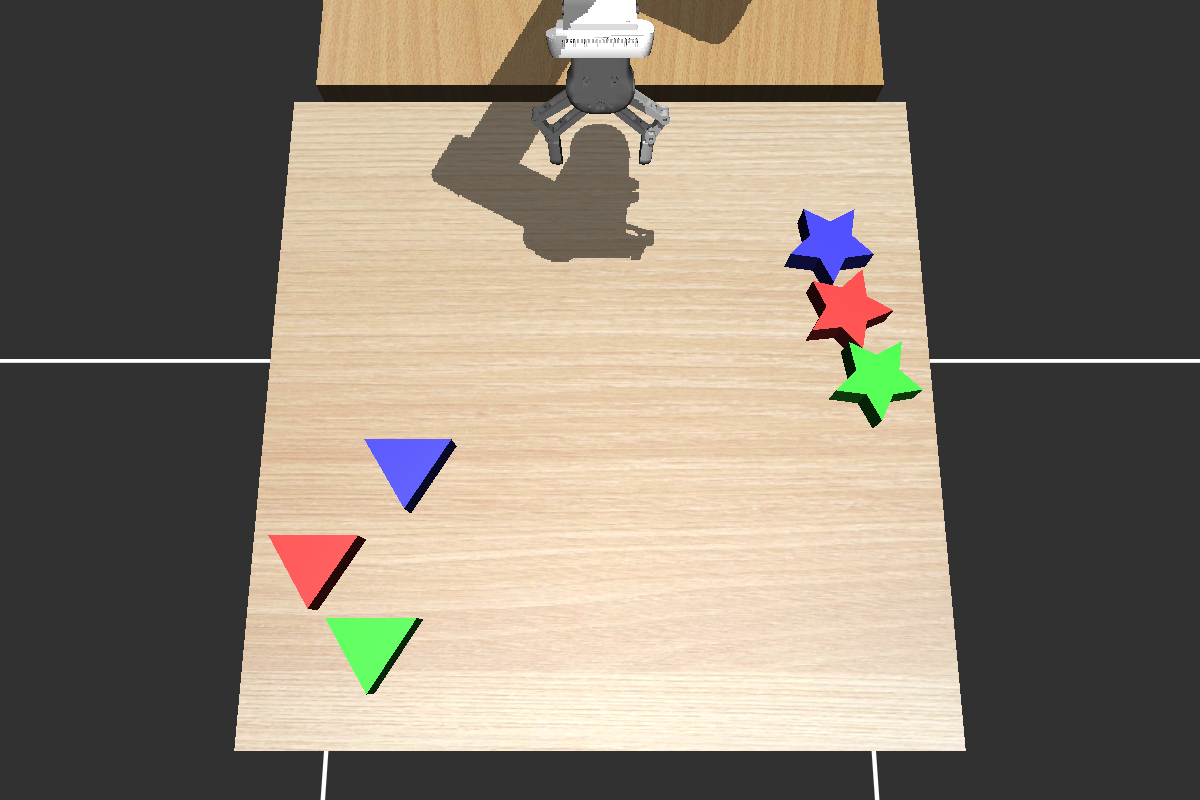}
			\label{block-preference}
		}
	\end{minipage}

    \caption{Examples of preferences for simulation scenario: (a) spatial pattern preferences arranged within a horizontal line, and (b) semantic preferences grouped with the same shaped objects.}
    \label{fig:simulation-preference-example}
\end{figure}

\begin{figure}[t]
    \centering
    \includegraphics[width=0.48\textwidth]{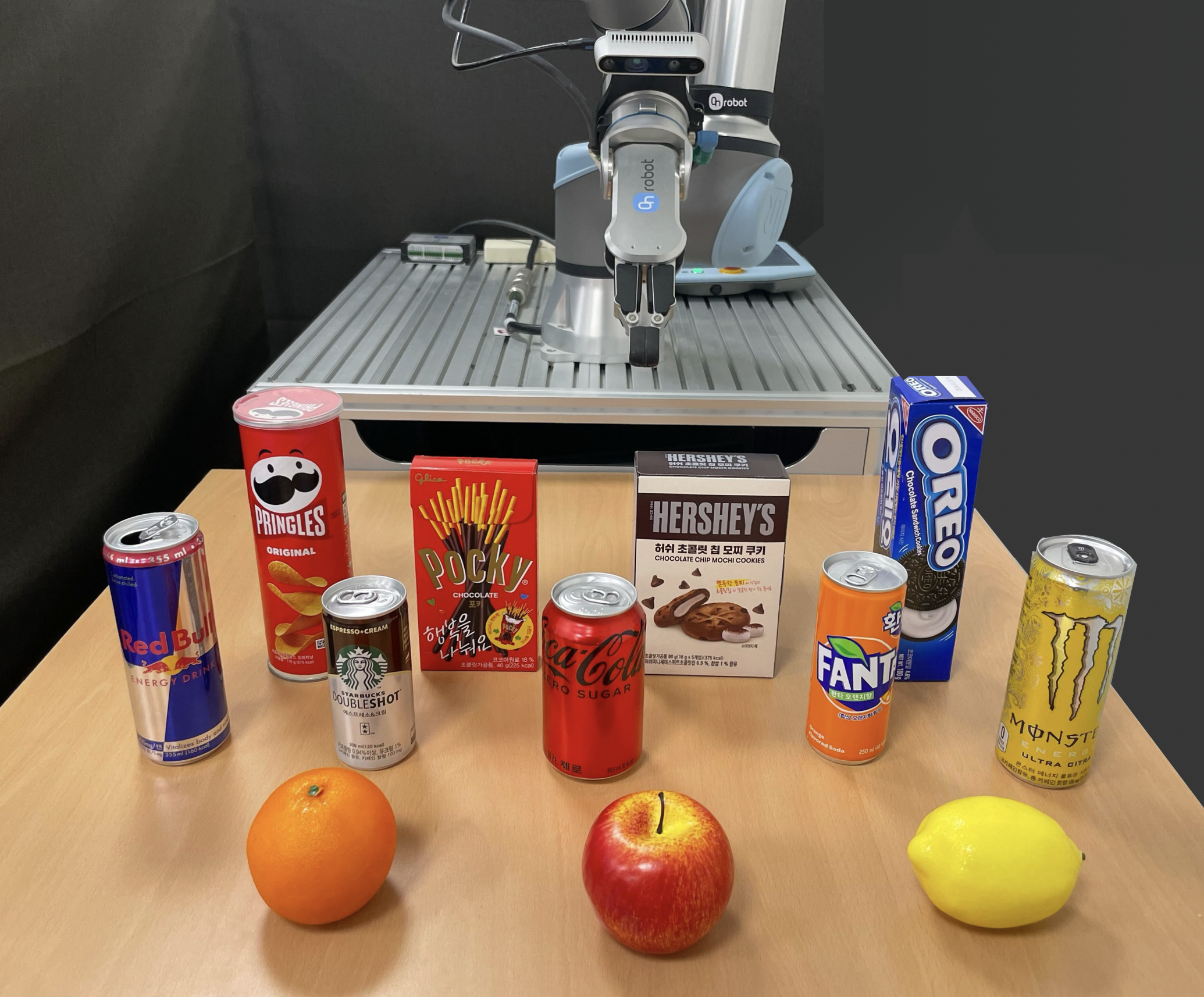}
    \caption{\textbf{Household objects:} We use various daily objects to test our approach, some of which can be categorized by terms of color, shape, or category.}
\label{fig:realworld-object}
\end{figure}

To validate the effectiveness of VRD, we conducted a set of experiments focusing on visual reasoning performance.
In addition, we designed further experiments to evaluate the ability to infer human preferences, which we divided into two categories: those based on semantic property and those based on spatial patterns.
Specifically, the performance of our proposed method is evaluated across three different environments: Block Task, Polygon Task, and Household Task.

First, we conduct experiments to demonstrate both the visual reasoning capability and the semantic preference reasoning capability in the Block Task where a robot moves blocks as shown in Fig.~\ref{blocks-only-preference}. 
For the Polygon Task in Fig.~\ref{block-preference}, we validate the performance of visual reasoning and spatial pattern preference reasoning with three triangular and three star-shaped polygons by rearranging the polygons. 
Finally, the Household Task is designed to demonstrate the applicability of our proposed method in real-world scenarios where various daily objects are used, as shown in Fig.~\ref{fig:realworld-object}.

\subsection{Baselines~\&~Metrics}
\label{baselines-metrics}
We compare our method with other baselines, including large language models and a linear preference extractor. For fair comparisons on visual reasoning, we utilize the same visual reasoning module (i.e., GPT-4V~\cite{00_GPT_4V}).

\begin{itemize}
    \item MLLM-Naive: An ablation of our approach that does not use the Visual Reasoning Descriptor and Preference Reasoning Descriptor.
    MLLM-Naive infers scene descriptions for consecutive image pairs in a similar way to our method but without using the VRD template.
    Then, this baseline interprets the preference directly, using only an entire image sequence in a single interaction.
    \item MLLM-L2R: Inspired by Language-to-Reward (L2R)~\cite{yu_23_language}, this baseline extracts normalized object 2D position (ranging from $0.0$ to $1.0$) information for feature computation. Subsequently, we integrate a code snippet generation module that produces a piece of code to compute preference weights using the obtained object positions.
    \item Mutual-Distance-based Preference Extractor~(MDPE): This baseline assumes that human preferences are deterministic, following a linear user model as discussed in prior works~\cite{wilde_18_pbl_linear,wilde_19_pbl_linear}. Within the framework of linear models, MDPE computes the preference weights for each specific feature based on pre-defined functions using the mutual distances between objects and then derives the preference from these weights.
\end{itemize}

\begin{table}[]
\centering
\footnotesize
\begin{tabular}{ccc}
\hline
\multicolumn{3}{c}{Siimulation Experiments: Block \& Polygon}                         \\ \hline
\multicolumn{1}{c|}{\multirow{2}{*}{Model}} & \multicolumn{2}{c}{$\text{SR}_{\text{VRD}}$~$\uparrow$}           \\ \cline{2-3} 
\multicolumn{1}{c|}{}                       & Block             & Polygon             \\ \hline
\multicolumn{1}{c|}{MLLM-CoVR (Ours)}       & \textbf{0.72$\pm$0.11} & \textbf{0.79$\pm$0.13} \\
\multicolumn{1}{c|}{MLLM-Naive}             & 0.40$\pm$0.15          & 0.56$\pm$0.23          \\ \hline
\end{tabular}
\caption{Table for visual reasoning experiments. The number (mean$\pm$standard deviation) indicates the success rate of predicting visual residuals in between images. A higher number indicates better performance.}
\label{exp:visual-reasoning}
\end{table}

Our evaluation metrics are the success rate of Visual Reasoning Descriptor ($\text{SR}_{\text{VRD}}$) and the success rate of Preference Reasoning Descriptor ($\text{SR}_{\text{PRD}}$) for the given image sequences. 
In particular, $\text{SR}_{\text{VRD}}$ is calculated based on the visual residual between the image sequences and is defined as follows:
\begin{equation*}
    \text{SR}_{\text{VRD}} = \frac{1}{N-1} \sum_{k=1}^{N-1} \left( \frac{\sum_{l \in V_k} \mathbb{I}(l = \hat{l})}{|V_k|} \right)
\end{equation*}
\noindent where \( \mid \cdot \mid \) means the number of elements in the set and \( \mathbb{I} \) represents an indicator function that checks whether each element \( l \) within the predicted response \( V_k \) matches its corresponding element \( \hat{l} \) in the ground truth visual residual \( \hat{V}_k \) for each consecutive image pair. 
The elements of \( V_k \) and \( \hat{V}_k \) include (\( l^{\text{semantic}}_k \), \( l^{\text{geometric}}_k \), \( l^{\text{description}}_k \)), and their respective ground truth counterparts (\( \hat{l}^{\text{semantic}}_k \), \( \hat{l}^{\text{geometric}}_k \), \( \hat{l}^{\text{description}}_k \)).

On the other hand, $\text{SR}_{\text{PRD}}$ is measured according to the predicted preference that matches the ground truth.
The preference criteria for each scene are manually designed.
We formulate $\text{SR}_{\text{PRD}}$ as follows:
\begin{equation*}
\text{SR}_{\text{PRD}} = \frac{1}{M} \sum_{i=1}^{M} \mathbb{I}(l^{\text{preference}}_{i} = \hat{l}^{\text{preference}}_{i})
\end{equation*}
where the indicator function $ \mathbb{I} $ checks for a match between predicted description and ground truth preferences. 
$\text{SR}_{\text{PRD}}$ evaluates whether the predicted preferences \( l^{\text{preference}}_{i} \) are in alignment with the ground truth preferences \( \hat{l}^{\text{preference}}_{i} \) across a defined set of scenes.

\subsection{Block Task: Spatial Pattern Preference Reasoning}
\paragraph{Setup}
In this task, we evaluate the performance of visual reasoning and preference reasoning from image sequences, assuming that only one object is moved in between sequential images.
This task contains six blocks of different colors that are rearranged in specific spatial patterns.
These patterns can be defined within specific regions of a tabletop, such as the right side, left side, or upper side of a table, or by forming specific patterns, such as vertical or horizontal alignments.
For example, as illustrated in Fig.~\ref{blocks-only-preference}, the blocks can be arranged to form a spatial pattern, such as a horizontal line along the center of the table.
We compared our approach with a naive model, performing 10 times for the spatial pattern preference set.

\paragraph{Results}
Firstly, Table \ref{exp:visual-reasoning} compares the visual reasoning performance of our method against the ablation of our method.
The results indicate that our method outperforms the naive approach in understanding both the semantic and geometric properties in between an image sequence.
Especially the results of 0.72$\pm$0.11 demonstrate the superior visual reasoning ability of our method.
In contrast, the MLLM-Naive model shows limited ability in extracting visual signals between images with $\text{SR}_{\text{VRD}}$ of 0.40$\pm$0.15 for the same task. 
This result highlights the effectiveness of our VRD template-based approach in recognizing the visual residuals within image sequences.

From the results in Table~\ref{exp:block-preference-reasoning}, we compare the spatial pattern preference reasoning performance of our method to three other baseline approaches.
Our method shows outstanding reasoning preference performance in the Block Task. 
This highlights the benefits of our prompting method as compared to the other baselines.
Although the MDPE approach was expected to get a perfect score, MDPE did not achieve a score of $1.0$ (indicating that this model is always correct). 
This poor performance of MDPE is mainly due to the parameter sensitivity used in its function function. 
Such sensitivity often leads to the recognition of multiple preferences, resulting in erroneous preference inferences.
MLLM-L2R also exhibits limited effectiveness primarily because relying solely on the responses of the MLLM for position information is impractical; they do not account for specific geometric locations.
From the experiment results, we would like to emphasize that our method has a high potential for visual reasoning tasks, taking into account spatial pattern preferences.

\begin{table}[]
\centering
\footnotesize
\begin{tabular}{cccc}
\hline
\multicolumn{4}{c}{Simulation Experiment: Block}                            \\ \hline
\multicolumn{1}{c|}{\multirow{2}{*}{Model}} & \multicolumn{3}{c}{$\text{SR}_{\text{PRD}}$~$\uparrow$} \\ \cline{2-4} 
\multicolumn{1}{c|}{}                       & quadrant    & vertical   & horizontal   \\ \hline
\multicolumn{1}{c|}{MLLM-CoVR (Ours)}       & 0.65        & \textbf{0.90}        & \textbf{0.80}          \\
\multicolumn{1}{c|}{MLLM-Naive}             & \textbf{0.85}        & 0.70       & 0.60          \\
\multicolumn{1}{c|}{MLLM-L2R}               & 0.25        & 0.70        & \textbf{0.80}          \\
\multicolumn{1}{c|}{MDPE}                   & 0.45        & 0.40        & 0.40          \\ \hline
\end{tabular}
\caption{\textbf{Block: Spatial Pattern Preference Reasoning.} Preferences are set for spatial patterns (i.e., quadrant, vertical, and horizontal alignments).}
\label{exp:block-preference-reasoning}
\end{table}

%
%
\subsection{Polygon Task: Semantic Preference Reasoning}
\paragraph{Setup}
We aim to show the performance of visual reasoning and semantic preference reasoning from image sequences in the Polygon Task.
This task consists of three triangular polygons and three star-shaped polygons. 
Our focus is on semantic preferences, specifically in grouping the objects based on either color or shape. The objective is to assess the ability to interpret semantic preferences derived from a sequence of images.
For example, as depicted in Fig.~\ref{block-preference}, objects are sorted based on the shape attributes.

\paragraph{Results} 
On the Polygon task, as detailed in Table \ref{exp:visual-reasoning}, our methodology significantly outperforms the baseline model, indicating superior visual reasoning accuracy.
Specifically, our approach achieves a visual reasoning accuracy of 0.79$\pm$0.13 while the baseline records a lower accuracy of 0.56$\pm$0.23.
This performance gap shows an enhanced ability of our method to accurately predict visual contexts within image sequences, especially in tasks involving complex geometric shapes such as polygons.

The results in Table~\ref{exp:polygon-preference-reasoning} show that our method consistently outperforms other MLLM-based approaches for semantic preference reasoning in the Polygon task.
In comparison, the MLLM-Naive method performs poorly with $0.20$ for color and $0.70$ for shape, indicating the limitations of naive models in capturing semantic properties.
On the other hand, the MLLM-L2R model shows slight improvements, achieving a score of $0.30$ for color and $0.80$ for shape.
However, these results still do not reach our reasoning preference performance.
Notably, MDPE achieves the perfect scores (=$1.0$) both on color and shape criteria. 
However, it is important to know that the performance of MDPE depends on the manual tuning of the preference feature functions.
These results imply that our method can successfully capture semantic preferences without any manual feature engineering, such as predefining a list of object attributes.  

\begin{table}[]
\centering
\footnotesize
\begin{tabular}{ccc}
\hline
\multicolumn{3}{c}{Simulation Experiment: Polygon}      \\ \hline
\multicolumn{1}{c|}{\multirow{2}{*}{Model}} & \multicolumn{2}{c}{$\text{SR}_{\text{PRD}}$ $\uparrow$} \\ \cline{2-3} 
\multicolumn{1}{c|}{}                 & color & shape \\ \hline
\multicolumn{1}{c|}{MLLM-CoVR (Ours)} & \textbf{1.0}   & 0.90  \\
\multicolumn{1}{c|}{MLLM-Naive}       & 0.20  & 0.70  \\
\multicolumn{1}{c|}{MLLM-L2R}         & 0.30  & 0.80  \\
\multicolumn{1}{c|}{MDPE}             & \textbf{1.0}   & \textbf{1.0}   \\ \hline
\end{tabular}%
\caption{\textbf{Polygon: Semantic Preference Reasoning.} Semantic (i.e., color and shape) preferences are evaluated. A score of $1.0$ is considered a perfect performance, indicating that the model consistently makes accurate predictions.}
\label{exp:polygon-preference-reasoning}
\end{table}

\begin{figure*}[h!]
    \centering
    \includegraphics[width=0.85\textwidth]{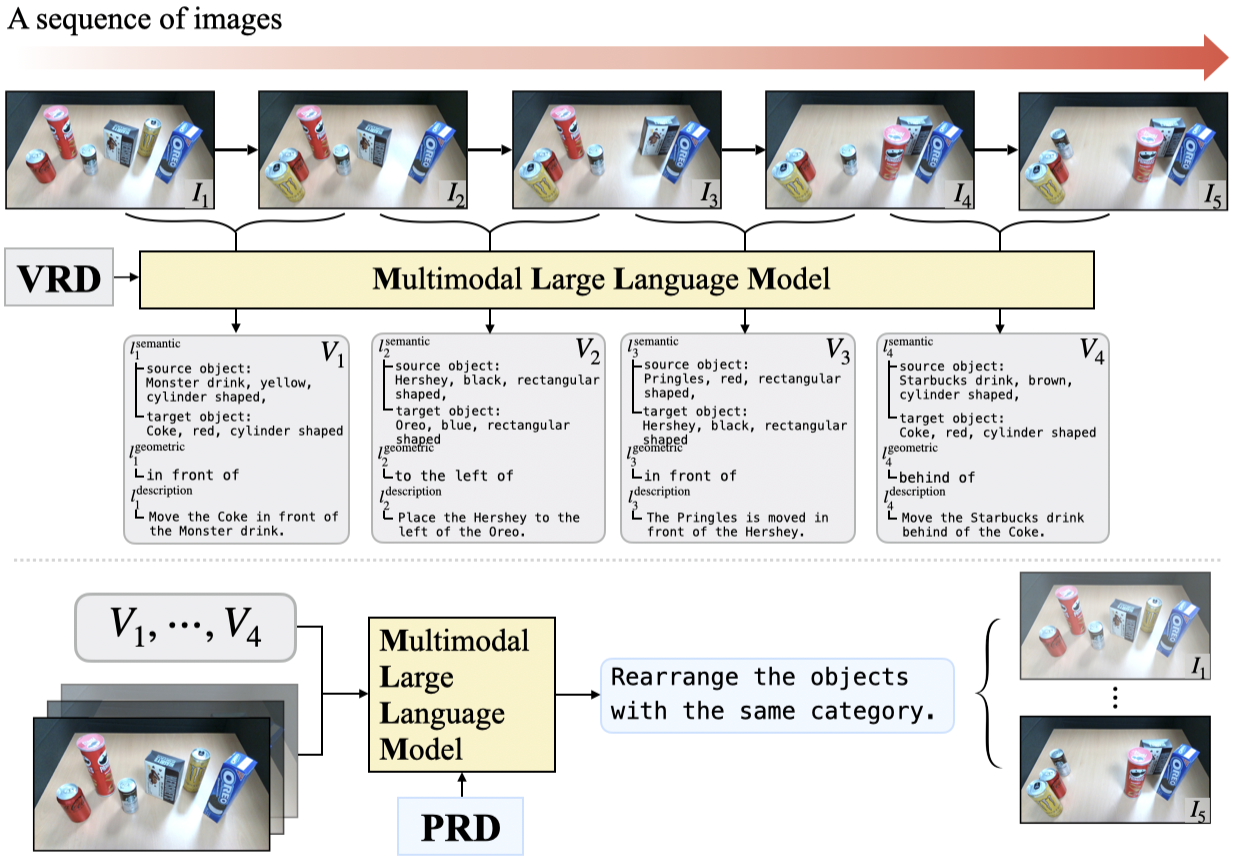}
    \caption{This figure illustrates the application of CoVR in a scenario where objects are rearranged based on their category. 
    The result of each visual residual shows the model's ability to identify semantic and geometric properties of objects, emphasizing the practical utility of CoVR in tasks that require a visual understanding of object properties and spatial relationships. 
    See more videos and tasks at \href{https://joonhyung-lee.github.io/vpi/}{https://joonhyung-lee.github.io/vpi/}}
\label{fig:run-example}
\end{figure*}

\begin{figure*}[t!] 
	\centering
	\subfigure[]{
	\includegraphics[width=0.292\linewidth]{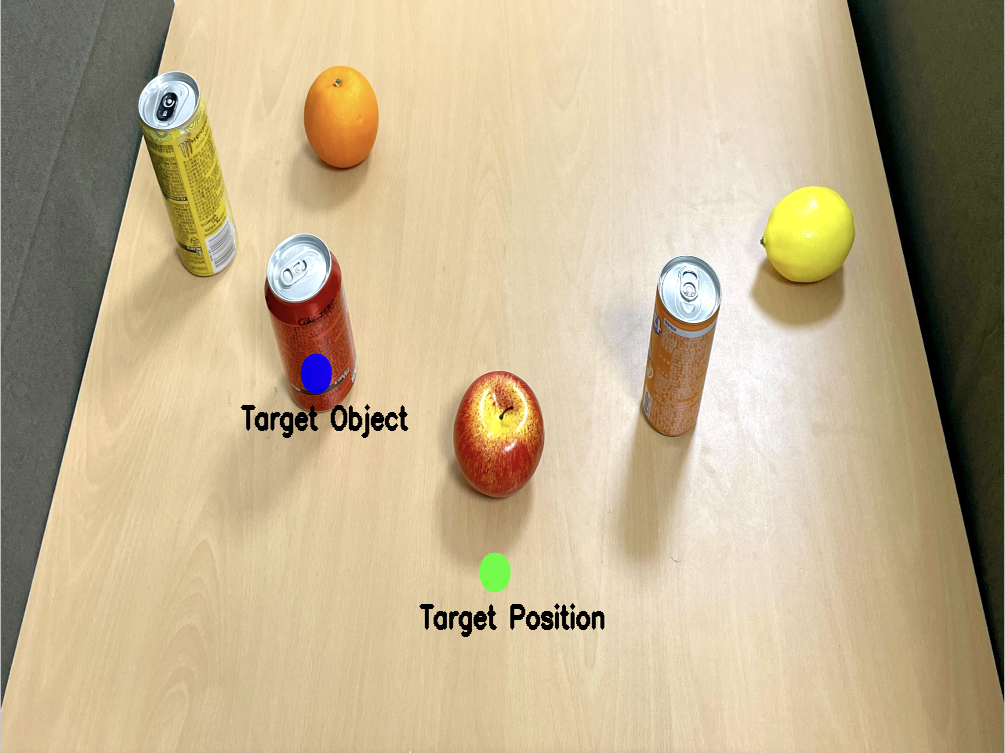}
	}
	\subfigure[]{
	\includegraphics[width=0.292\linewidth]{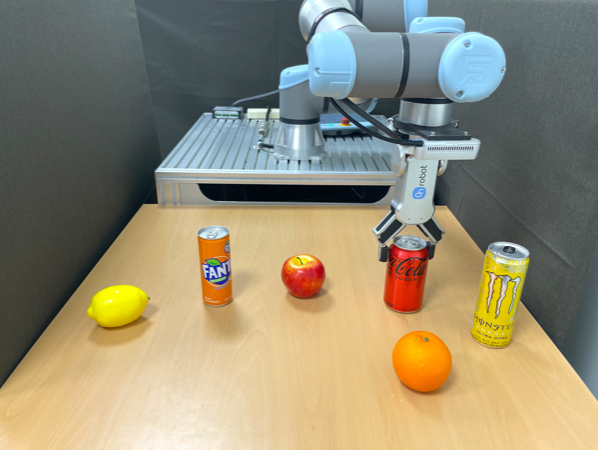}
	}
	\subfigure[]{
	\includegraphics[width=0.292\linewidth]{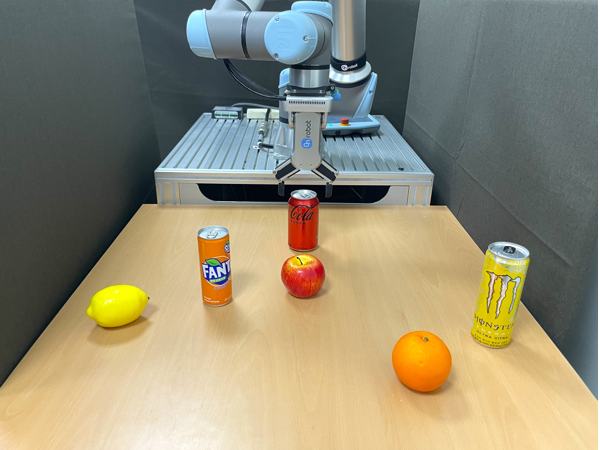}
	}
	\subfigure[]{
	\includegraphics[width=0.292\linewidth]{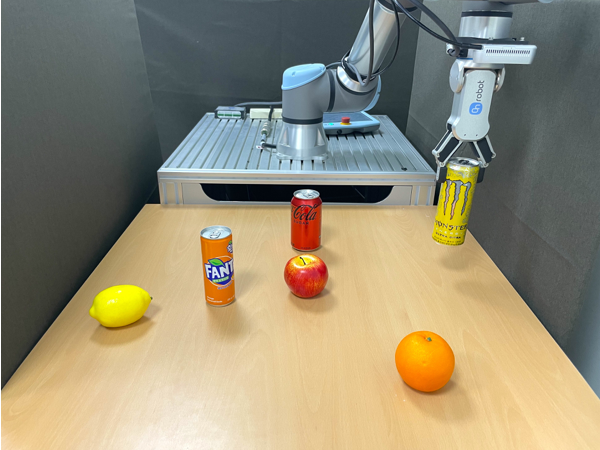}
	}
	\subfigure[]{
	\includegraphics[width=0.292\linewidth]{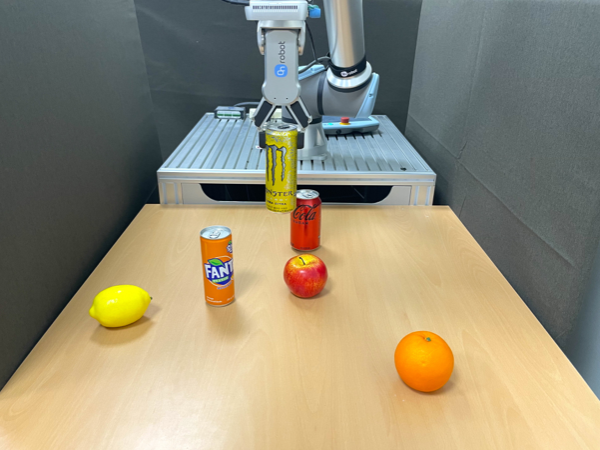}
	}
	\subfigure[]{
	\includegraphics[width=0.292\linewidth]{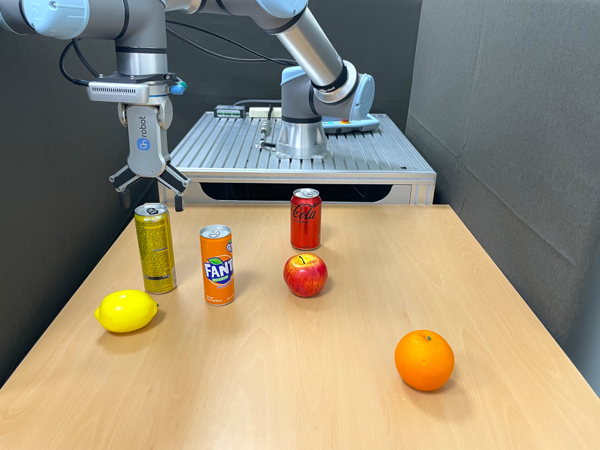}
	}
	\subfigure[]{
	\includegraphics[width=0.292\linewidth]{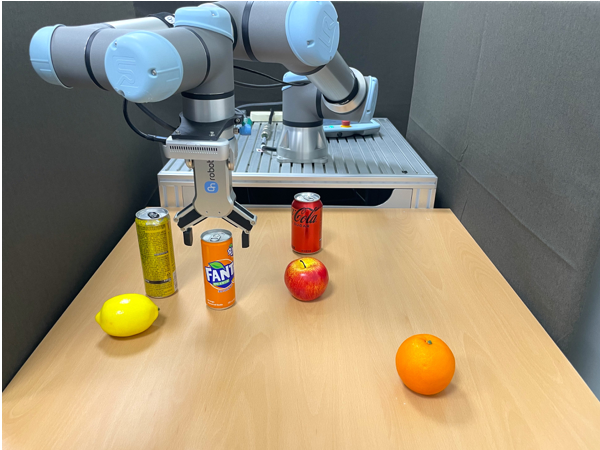}
	}
	\subfigure[]{
	\includegraphics[width=0.292\linewidth]{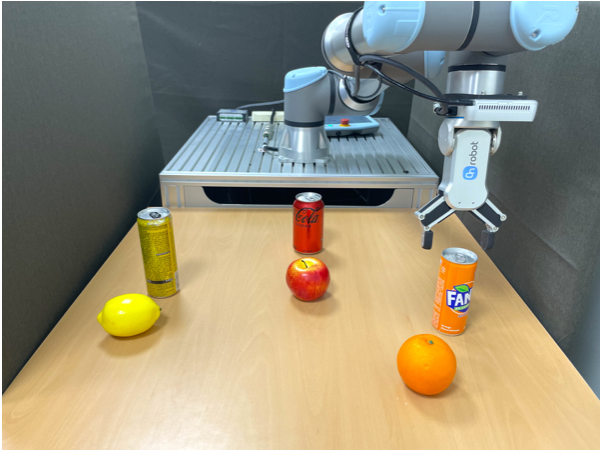}
	}
	\subfigure[]{
	\includegraphics[width=0.292\linewidth]{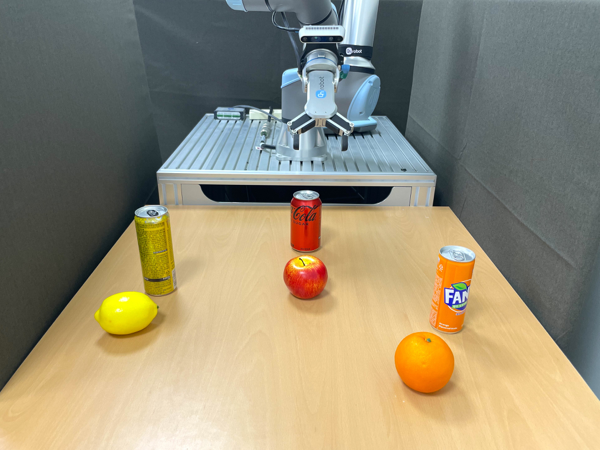}
	}         
	\caption{\textbf{Real-world demonstration snapshot.} 
    This snapshot illustrates the application of semantic preference in object rearrangement based on color matching, which is demonstrated through a series of robot actions: (a) shows the user interface as perceived by the user, where the blue circle represents the object to be moved and the green circle represents the desired coordinates, (b)-(c) show the robot placing a red drink next to an apple to align by color, (d)-(f) depict the robot placing a yellow drink in front of a lemon, and (g)-(i) finally show the robot placing an orange drink close to an orange.
    We demonstrate our method on real robot hardware, with hard-coded procedures for picking and placing objects.}
	\label{fig:realworld-snapshot}
\end{figure*}

\subsection{Household Task: Real-world Demonstration}
\paragraph{Setup} The Household Task includes three object types: Fruits, Snacks, and Beverages.
This task focuses on placing objects based on the semantic properties or within the spatial patterns, including preferences for both semantic and spatial arrangements.
We use 12 real-world testing objects, as shown in Fig. \ref{fig:realworld-object}, to evaluate the ability to identify both the semantic and the spatial preferences.
We evaluate our approach in real-world tabletop environments with a 6-DoF UR5e manipulator with an OnRobot RG2 gripper. 

\paragraph{Results} 
The metric of $\text{SR}_{\text{VRD}}$, presented in Table \ref{exp:real-world-demonstration} compared the visual reasoning performance of our approach against the ablation of our method, which was evaluated six times respectively.
In particular, the results of 0.63$\pm$0.08 demonstrated the higher visual reasoning performance of our method.
In contrast, the MLLM-Naive model showed a limited ability to extract visual signals between images with $\text{SR}_{\text{VRD}}$ of 0.28$\pm$0.19 for the same task. 
These results support the effectiveness of our VRD template-based approach in recognition of visual residuals within image sequences.

Each type of preference was evaluated six times, and performance was measured in terms of $\text{SR}_{\text{PRD}}$.
As illustrated in Fig.~\ref{fig:realworld-snapshot}, in each step, the robot performs to move objects and captures images.
The results of our method in Table \ref{exp:real-world-demonstration} are consistent with our simulation results, indicating the balanced performance of our method in spatial pattern and semantic preference reasoning. 
Compared to other MLLM-based approaches, they showed subpar performance in recognizing spatial patterns and semantic properties. 
We can notice that MLLM tends to misunderstand the spatial arrangements or semantic properties of objects without explicit annotation by VRD.
While MDPE performs as effectively as our approach for both types of preference, it remains highly dependent on the need for handcrafted features.
These results support the practical effectiveness of our method and support its successful application in real-world scenarios.


\begin{table}[]
\centering
\footnotesize
\begin{tabular}{cccc}
\hline
\multicolumn{4}{c}{Real-world Experiment: Household}                                                                         \\ \hline
\multicolumn{1}{c|}{\multirow{2}{*}{Model}} & \multicolumn{1}{c|}{\multirow{2}{*}{$\text{SR}_{\text{VRD}}$~$\uparrow$}} & \multicolumn{2}{c}{$\text{SR}_{\text{PRD}}$~$\uparrow$}      \\ \cline{3-4} 
\multicolumn{1}{c|}{}                       & \multicolumn{1}{c|}{}                        & Spatial Pattern & Semantic      \\ \hline
\multicolumn{1}{c|}{MLLM-CoVR (Ours)}       & \multicolumn{1}{c|}{\textbf{0.63$\pm$0.08}}                        & \textbf{0.67}   & \textbf{0.67} \\
\multicolumn{1}{c|}{MLLM-Naive}             & \multicolumn{1}{c|}{0.28$\pm$0.19}                        & 0.17            & 0.33          \\
\multicolumn{1}{c|}{MLLM-L2R}               & \multicolumn{1}{c|}{-}                       & 0.17            & 0.33          \\
\multicolumn{1}{c|}{MDPE}                   & \multicolumn{1}{c|}{-}                       & 0.50            & \textbf{0.67} \\ \hline
\end{tabular}
\caption{\textbf{Household: Real-world demonstration.} Each preference type is evaluated six times for both spatial pattern preference and semantic preference.}
\label{exp:real-world-demonstration}
\end{table}

\subsection{Limitation}
Although our proposed method shows its effectiveness in multiple scenarios, our method is heavily reliant on the image sequence, which makes it highly dependent on perception and the order of images. 
For instance, a cylinder-shaped beverage might appear as a round object depending on the viewpoint of the camera.
Therefore, our method may generate inaccurate scene descriptions, such as mismatching properties of objects or misunderstanding spatial relationships.
We believe that one way to address this issue is to facilitate MLLM's understanding of human correction by incorporating human feedback.

%
%
\section{Conclusions}
In this paper, we introduce a Visual Preference Inference (VPI) task designed to infer user preferences using visual reasoning from a series of images in the context of tabletop object manipulation.
We have demonstrated the effectiveness of our method in interpreting spatial relations from image sequences and inferring preferences in both simulation and real-world tabletop environments.
While our approach shows potential performance, it is critical to recognize its reliability on image sequences. Such dependence may result in sensitivity to perceptual biases and the order of images, which can lead to inaccuracies.
Despite these limitations, our work presents a significant step toward enhancing the ability to understand preferences in manipulation tasks, opening avenues for further research in the field of reasoning preferences in the robotics domain.



\bibliographystyle{IEEEtran}
\bibliography{references}  

\end{document}